\documentclass[10pt,twocolumn,letterpaper]{article}

\usepackage{cvpr}              
\usepackage[accsupp]{axessibility}  
\definecolor{cvprblue}{rgb}{0.21,0.49,0.74}
\usepackage[pagebackref,breaklinks,colorlinks,allcolors=cvprblue]{hyperref}

\usepackage{xcolor}
\definecolor{lzcolor}{RGB}{0, 102, 204}

\def\paperID{6043} 
\def\confName{CVPR}
\def\confYear{2026}

\title{Masked Next-Scale Prediction for Self-supervised Scene Text Recognition}

\author{
    Zhuohao Chen\textsuperscript{1} \quad
    Zeng Li\textsuperscript{2} \quad
    Yifei Zhang\textsuperscript{2} \quad
    Chang Liu\textsuperscript{3}\thanks{Corresponding authors.} \quad
    Yu Zhou\textsuperscript{1}\footnotemark[1] \\
    {\small \textsuperscript{1}Nankai University}\\
    {\small \textsuperscript{2}Institute of Information Engineering, Chinese Academy of Sciences}\\
    {\small \textsuperscript{3}Department of Automation and BNRist, Tsinghua University}\\
    {\tt\small chenzhuohao1120@gmail.com, liuchang2022@tsinghua.edu.cn, yzhou@nankai.edu.cn}
}

\begin{document}

\maketitle

\begin{abstract}
Scene Text Recognition requires modeling visual structures that evolve from coarse layouts to fine-grained character strokes. 
Training such models relies on large amounts of annotated data. 
Recent self-supervised approaches, such as Masked Image Modeling (MIM), alleviate this dependency by leveraging large-scale unlabeled data. 
Yet most existing MIM methods operate at a single spatial scale and fail to capture the hierarchical nature of scene text.
In this work, we introduce Masked Next-Scale Prediction (MNSP), a unified self-supervised framework designed to explicitly model cross-scale structural evolution. 
The framework incorporates Next-Scale Prediction (NSP), which learns hierarchical representations by predicting higher-resolution features from lower-resolution contexts. 
Naive scale prediction, however, tends to produce spatially diffuse attention, directing the model toward background regions rather than textual structures.
MNSP resolves this limitation by jointly learning cross-scale prediction and masked image reconstruction. 
NSP captures global layout priors across resolutions, while masked reconstruction imposes strong local constraints that guide attention toward informative text regions. 
A Multi-scale Linguistic Alignment module further maintains semantic consistency across different resolutions.
Extensive experiments demonstrate that MNSP achieves state-of-the-art performance, reaching 86.2\% average accuracy on the challenging Union14M benchmark and 96.7\% across six standard datasets. 
Additional analyses show that our method improves robustness under extreme scale and layout variations. Code is available at \url{https://github.com/CzhczhcHczh/MNSP}
\end{abstract}    
\section{Introduction}
\label{sec:intro}

\begin{figure*}[!t]
\centering
\includegraphics[width=0.99\linewidth]{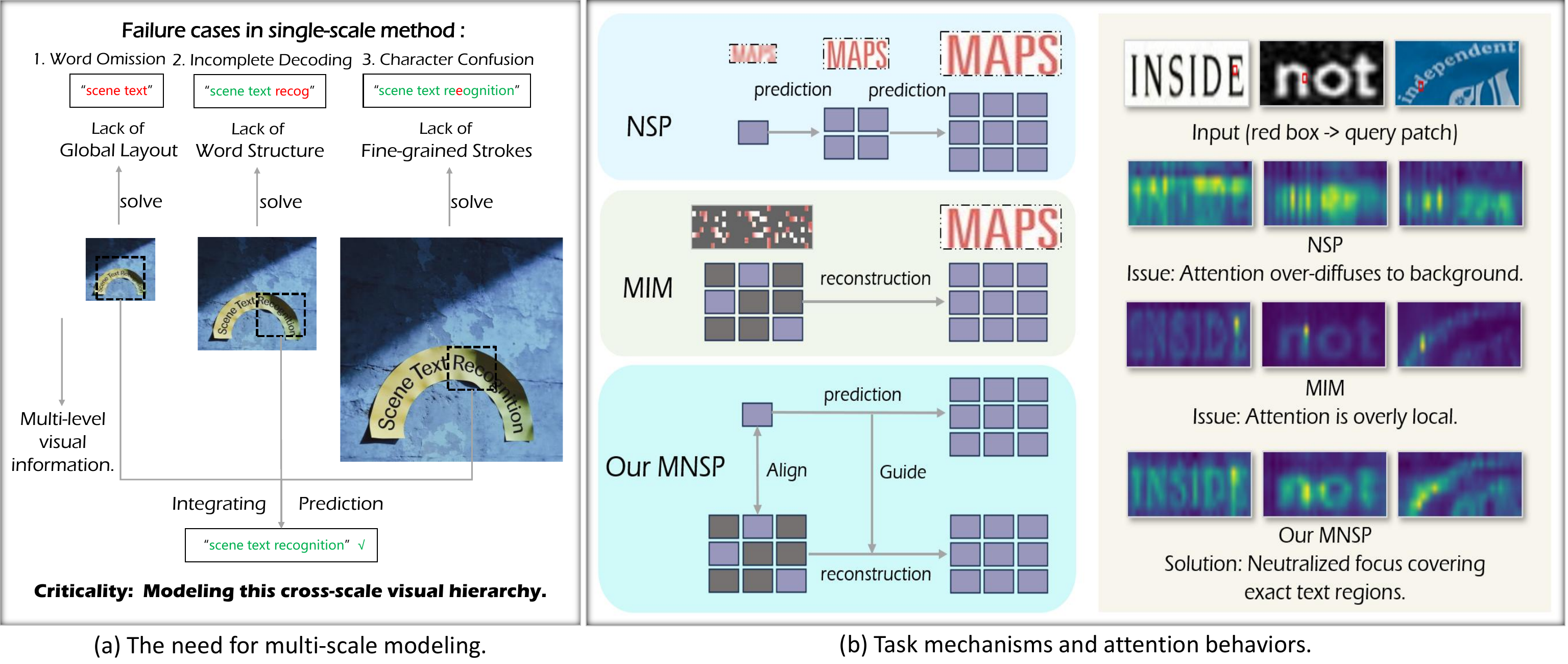}  
\caption{Overview of our motivation and core mechanism.
(a) STR inherently requires interpreting a multi-granular visual hierarchy. Single-scale features fail to simultaneously capture global layouts, word structures, and fine-grained strokes, which inevitably leads to word omissions, incomplete decoding, or character confusion, respectively. 
(b) We reveal the mechanism duality of two paradigms. NSP provides global structural priors but suffers from attention diffusion (top). MIM provides local fine-grained priors but is too short-sighted (middle). By deeply coupling them via alignment and guidance, our MNSP successfully neutralizes their defects, achieving coherent and precise focus over the entire text region (bottom).}
\label{fig:fig1}  
\vspace{-10pt}
\end{figure*}

Scene Text Recognition (STR) is a fundamental vision–language task that aims to transcribe text from natural scene images. It serves as a critical component for many real-world applications~\cite{shu_arxiv_2025, yang_ijcv_2025}, including document digitization, assistive reading, and visual understanding systems. Despite remarkable progress in deep learning–based methods~\cite{PerturbCTC_lizeng_STR, pimnet_str}, STR remains challenging due to the intrinsic complexity of scene text, which often exhibits severe appearance variations such as diverse fonts, irregular layouts, perspective distortions, and complex backgrounds.

Beyond these challenges, an often overlooked but fundamental property of STR is its hierarchical visual nature. Reading text inherently requires perceiving visual information across multiple scales: a model must first capture the global layout to locate text lines, then understand intermediate word-level structures, and finally distinguish fine-grained character strokes to resolve visually ambiguous symbols (Fig.~\ref{fig:fig1}a ). This hierarchical perception suggests that effective STR representations should explicitly encode cross-scale dependencies.

To alleviate the heavy reliance on annotated data, self-supervised learning (SSL) has been widely explored. The paradigm of SSL has evolved significantly, transitioning from early handcrafted pretext tasks~\cite{context_iccv15,yaoyuan_cvpr_20, dezhao_aaai_20, liu_tnnls} and contrastive learning approaches~\cite{he_moco_cvpr20,zhang_icpr_2020, li_mm_2021, fang_mm_2022, zhang_tmm_2024} to the currently Masked Image Modeling (MIM)~\cite{mae_cvpr22}. Inspired by these general advancements, MIM has also emerged as a prevailing framework for scene text recognition (STR), where models learn visual representations by reconstructing masked patches. Several works~\cite{maskocr_tmlr24, qiao2023decoupling, zhang2025lmim} extend this framework by incorporating linguistic priors, specialized masking strategies, or hybrid generative–contrastive objectives. While these approaches demonstrate strong representation learning capabilities, they share a common limitation: they operate at a single spatial scale. Consequently, existing SSL methods fail to explicitly model the hierarchical cross-scale structure that is fundamental to scene text perception.

To address this limitation, we explore Next-Scale Prediction (NSP) as a mechanism for learning hierarchical visual representations. NSP predicts higher-resolution features from lower-resolution contexts, naturally modeling the coarse-to-fine evolution of visual structures. Such a paradigm aligns well with how text is visually formed and perceived. However, directly applying NSP to STR introduces an unexpected challenge. Because all tokens remain globally visible during prediction, the encoder’s attention tends to become spatially diffuse, causing the model to attend to irrelevant background regions rather than focusing on text structures, as shown in Fig.~\ref{fig:fig1}b.

Notably, we find that MIM exhibits an inductive bias that naturally complements this behavior. While NSP captures global structural priors across scales, MIM enforces strong local constraints: masked tokens must rely on nearby visible patches to reconstruct missing visual details. As a result, MIM encourages spatially concentrated attention on informative regions. This complementary property suggests that combining the two objectives could simultaneously enable global cross-scale reasoning and precise local perception.

Motivated by this insight, we propose Masked Next-Scale Prediction (MNSP), a unified self-supervised framework that integrates cross-scale prediction with masked reconstruction. In this framework, NSP explicitly models hierarchical visual structures, while MIM suppresses attention diffusion and improves spatial precision.

However, learning representations across multiple resolutions may introduce semantic inconsistencies, where features extracted at different scales encode mismatched textual semantics. To address this issue, we further introduce a Multi-scale Linguistic Alignment (MLA) module. MLA explicitly aligns semantic representations across different scales and views, ensuring that hierarchical visual features remain linguistically consistent during representation learning. By enforcing this cross-scale semantic agreement, MLA stabilizes the joint optimization of NSP and MIM and improves the quality of learned representations.

We validate MNSP on both the Union14M benchmark~\cite{union14m_sstr_iccv23} and six common STR benchmarks. With a ViT-S backbone, MNSP achieves 86.2\% on Union14M and 96.7\% averaged over six standard datasets. Our controlled comparisons against single-scale baselines further verify that explicit multi-scale modeling brings significantly stronger robustness to extreme scale variations. 

Our main contributions are summarized as follows:
\begin{itemize}
    \item We explicitly target the inherently multi-scale challenge of STR by introducing \textbf{Next-Scale Prediction} into self-supervised representation learning, enabling the encoder to capture the hierarchical evolution from coarse layouts to fine strokes.
    \item We identify the attention diffusion defect of the NSP objective and solve it by introducing Masked Image Modeling as a complementary constraint. This dual-paradigm coupling successfully neutralizes attention, perfectly merging global structural priors with local fine-grained precision.
    \item We design a tightly coupled architecture utilizing NSP Guidance and a \textbf{Multi-scale Linguistic Alignment} module. This ensures that cross-scale structural learning and masked reconstruction maintain strict semantic consistency within a single forward pass.
    \item Experiments demonstrate that MNSP achieves top accuracy on the Union14M benchmark (86.2\%) and six standard STR benchmarks (96.7\%), exhibiting exceptional robustness against severe scale and layout variations.
\end{itemize}

\section{Related Work}
\label{sec:relatedwork}
\textbf{Supervised Scene Text Recognition.}
Supervised STR methods typically adopt specialized decoding strategies to align visual features with textual sequences, including CTC-based models~\cite{crnn,du2022svtr,du2025svtrv2}, autoregressive decoders~\cite{qiao2020seed,abinet_cvpr21}, and parallel attention-based frameworks~\cite{srn_cvpr20,na2022matrn}. 
To address scale and layout variations in natural scenes, several approaches incorporate multi-scale representations through feature pyramids or scale-aware attention mechanisms, such as SAFE~\cite{SAFE_accv2018}, SaHAN~\cite{SaHAN_PR}, TPS++~\cite{TPS++_ijcai_2023}, and ANT-STR~\cite{ANT-STR_2023}. However, these multi-scale designs are fundamentally \textit{implicit}—they fuse features across resolutions to maintain scale invariance without explicitly supervising the cross-scale structural mapping. 
Recent works further enhance recognition by integrating stronger semantic modeling or pre-trained language models~\cite{wang2023symmetrical,fujitake2024dtrocr}. 
Despite their effectiveness, these methods rely heavily on large-scale annotated datasets, which limits their scalability and motivates the exploration of self-supervised learning.

\textbf{Self-supervised Scene Text Recognition.}
To reduce the dependence on labeled data, recent studies exploits the intrinsic visual and semantic regularities within large-scale unlabeled datasets. Early methods adopt contrastive learning to align augmented views of images~\cite{seqclr,ccd_sstr_iccv23}. Recent approaches are predominantly based on MIM, where models learn visual representations by reconstructing masked patches. Approaches like MAERec~\cite{union14m_sstr_iccv23} reconstruct masked text patches to capture fine-grained stroke details, while advanced variants such as Dual-MAE~\cite{qiao2023decoupling}, DiG~\cite{dig_sstr_mm22}, and LMIM~\cite{zhang2025lmim} attempt to decouple visual and semantic features or inject linguistic priors to enhance representation robustness.
Despite their effectiveness, most existing SSL approaches operate at a fixed spatial scale and mainly capture local textures but completely bypass the hierarchical, cross-scale structural dependencies that define text composition. The joint integration of multi-scale structure and local masked reconstruction remains underexplored.

\textbf{Generative Next-Scale Prediction.}
NSP has recently emerged as an effective paradigm for hierarchical visual generation, notably popularized by VAR~\cite{var} and its extensions~\cite{mar,tang2024hart,varsr}. These models formulate image synthesis as a coarse-to-fine autoregressive process, explicitly predicting higher-resolution token maps conditioned on lower-resolution ones. While such global-to-local structural modeling naturally reflects the hierarchical nature of visual content, existing NSP frameworks are primarily designed for generative modeling and rely on discrete token codebooks. Consequently, their direct application to self-supervised representation learning remains largely unexplored. To overcome these architectural barriers, we adapt NSP into a codebook-free, end-to-end latent mapping via one-step prediction.
\section{Method}
\label{sec:method}

\begin{figure*}[!t]
\centering
\includegraphics[width=0.99\linewidth]{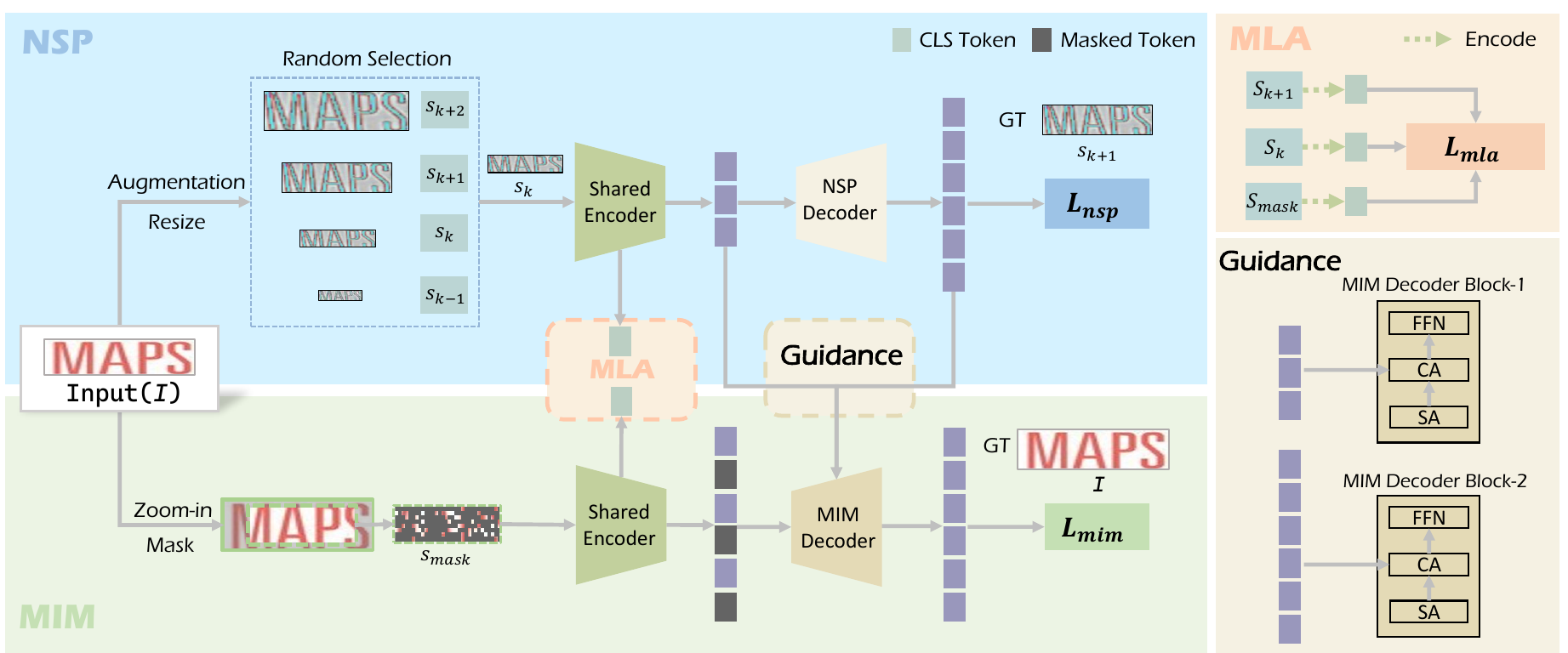}  
\caption{Architecture overview of our MNSP framework. The model consists of three tightly coupled modules sharing an online encoder.
(1) NSP Branch: Randomly selects adjacent scales ($s_k, s_{k+1}$) and performs explicit next-scale latent prediction to capture global structural evolution.
(2) MIM Branch: Reconstructs masked local tokens from a zoomed-in view ($s_{mask}$). Crucially, it operates under the Guidance of NSP: the Cross-Attention (CA) layers in MIM decoders sequentially inject layout cues ($\mathcal{E}(s_k)$) and fine-grained predictions ($D_{nsp}(F)$) from the NSP branch, grounding global priors into local details.
(3) MLA Module: Multi-scale Linguistic Alignment explicitly aligns the [CLS] tokens across diverse views ($s_k, s_{k+1}, s_{mask}$) to guarantee scale-invariant semantic consistency within a single forward pass.}
\label{fig:framework}  
\vspace{-10pt}
\end{figure*}

\textbf{Overall Architecture.}
The overall architecture of our framework is shown in Fig.~\ref{fig:framework}. MNSP performs scale-aware self-supervision through three components:
(1) an NSP branch that selects a scale pair from a 4× adjacency sequence and performs direct scale-to-scale prediction as the main multi-scale learner, strengthening the shared encoder without codebooks;
(2) a lightweight MIM branch that reconstructs masked regions using NSP-guided contextual cues to enhance local perception; and
(3) a multi-scale linguistic alignment module that provides appearance-invariant text cues, guides MIM reconstruction, and enforces cross-scale semantic consistency.
Together, these components enable efficient, explicit, and linguistically aligned multi-scale learning within a single unified pipeline.

\subsection{Next-Scale Prediction}
The core of MNSP lies in its NSP branch, which reformulates multi-scale representation learning as a direct scale-to-scale prediction problem. Instead of relying on codebooks or quantized latent tokens as in VQ-based methods, NSP learns a continuous mapping from smaller-scale features to their larger-scale counterparts.

Giving an adjacent scale pair $s_k, s_{k+1}$, the encoder output at the smaller scale $\mathcal{E}(s_k)$, is bicubically upsampled to $F$ and fed into the NSP decoder. The decoder then predicts the higher-resolution feature map $\mathcal{D}_{nsp}(F)$ under the supervision of the frozen teacher encoder $\mathcal{E}_t (s_k{+1})$. This teacher-student setting stabilizes training and provides hierarchical feature targets derived from MAERec-pretrained representations, ensuring that NSP captures both semantic continuity and visual fidelity across scales.

Different from pixel-level upsampling, NSP operates entirely in the latent space, enabling the model to explicitly learn inter-scale correspondences such as character shape consistency and text-line layout proportionality. By enforcing the prediction of a larger-scale representation directly from a smaller-scale embedding, NSP acts as a scale bridge, transferring structural regularities upward while retaining the linguistic invariance inherited from augmentation diversity. This mechanism transforms the self-supervised objective from intra-scale reconstruction into cross-scale feature alignment, effectively teaching the encoder to reason over text at multiple visual granularities simultaneously.

We adopt a patch size $p{=}4$ and a geometric sequence of text-specific resolutions ${s_1,s_2,s_3,s_4}$ with adjacent scales satisfying $s_{k+1}{=}(2h,2w)$ if $s_k{=}(h,w)$. The number of patch tokens is $N(s){=}\tfrac{h}{p}\tfrac{w}{p}$, hence $N(s_{k+1}){=}4N(s_k)$. At training time we uniformly sample one adjacent pair $(s_k,s_{k+1})$. Let $F\in\mathbb{R}^{N(s_{k+1})\times d}$ denote the bicubic upsampled tokens from $\mathcal{E}(s_k)$ (the patch tokens; \texttt{[CLS]} is excluded). Each NSP decoder block follows an SA–CA–FFN routine with the standard attention
\begin{equation}
A(Q,K,V)=\mathrm{Softmax}\Big(\tfrac{QK^\top}{\sqrt{d}}\Big)V,
\end{equation}
where self-attention uses $(Q,K,V){=}(F,F,F)$, and cross-attention takes $K,V$ from the small-scale encoder features, i.e., $(Q,K,V){=}(F,\mathcal{E}(s_k),\mathcal{E}(s_k))$. The NSP loss is a feature MSE against the teacher target (excluding \texttt{[CLS]}):
\begin{equation}
\mathcal{L}_{\text{nsp}}
=\frac{1}{N(s_{k+1})}||\mathcal{D}_{\text{nsp}}(F)-\mathcal{E}_t(s_{k+1})||_2^2.
\label{eq:nsp}
\end{equation}

The scale-to-scale prediction objective implicitly strengthens fine-grained representation learning at smaller scales. 
Because the higher-resolution feature map contains richer spatial details, predicting it from a lower-resolution embedding requires the encoder to preserve sufficient local information in the small-scale representation. 
As a result, the encoder is encouraged to encode detailed character strokes and local text structures even under coarse inputs, improving its ability to capture fine-grained visual cues.

\subsection{Masked Image Modeling with NSP Guidance}

NSP equips the model with global cross-scale structure, while the MIM branch is responsible for capturing micro-level cues that determine text appearance.
However, conventional MIM reconstructs masked regions using only intra-image context, which often leads to locally plausible but semantically isolated predictions.
In MNSP, we instead let MIM decode under explicit guidance from the NSP branch, so that local completion is conditioned on both the layout at a smaller scale and the structure predicted at a larger scale (see Fig.~\ref{fig:framework}).

The MIM branch takes a zoomed-in and randomly masked view of the input image $I$.
The shared encoder processes only the visible patches of this view and outputs their embeddings, which are then combined with mask tokens to form the full sequence $F_{\text{mim}}$ fed into a two-block decoder.

In the first block, cross-attention is applied between $F_{\text{mim}}$ and the encoded tokens of the small-scale image $s_k$ (excluding the [CLS] token), \emph{i.e.}, we set $(Q,K,V)=(F_{\text{mim}},\mathcal{E}(s_k),\mathcal{E}(s_k))$ in Eq.~(1).
This injects layout-level cues such as word-line geometry and inter-character spacing, anchoring each partially observed region in a coherent global structure.
In the second block, we reuse the same attention form but replace the keys and values with the NSP prediction at the next scale, $(Q,K,V)=(F_{\text{mim}},\mathcal{D}_{\text{nsp}}(F),\mathcal{D}_{\text{nsp}}(F))$.
These decoded features encode how strokes evolve from $s_k$ to $s_{k+1}$, providing a stroke-level prior that sharpens contours and enforces continuity when reconstructing masked patches.

This two-level guidance---layout from $\mathcal{E}(s_k)$ and strokes from $\mathcal{D}_{\text{nsp}}(F)$---closely couples MIM and NSP while keeping decoders specialized.
Gradients from the MIM loss flow not only to the shared encoder but also into $\mathcal{D}_{\text{nsp}}$ through the second block, reinforcing scale-to-scale consistency.

Let the MIM decoder output be $\mathcal{D}_{\text{mim}}(F_{\text{mim}})$ and the target come from the frozen target encoder on the original image, $\mathcal{E}_t(I)$.
With a binary mask $m_i\in\{0,1\}$ indicating masked tokens, the MIM loss is defined as
\begin{equation}
\mathcal{L}_{\text{mim}}
=\frac{1}{\sum_i m_i}\sum_i m_i
\left\|
\mathcal{D}_{\text{mim}}(F_{\text{mim}})_i
-
\mathcal{E}_t(I)_i
\right\|_2^2.
\label{eq:mim}
\end{equation}
NSP focuses on learning stable cross-scale structure, whereas MIM emphasizes local completeness under occlusion; using separate decoders prevents gradient conflicts between these objectives, while the shared encoder enforces a common multi-scale representation space.

\subsection{Multi-scale Linguistic Alignment}
While NSP and MIM respectively enhance cross-scale structural reasoning and fine-grained local perception, they operate primarily in the visual feature space.
However, self-supervised learning for STR requires more than visual consistency: the textual semantics encoded in the visual patterns must remain invariant across different scales, augmentations, and masking patterns. Without explicit semantic alignment, the shared encoder may develop scale-specific semantic drift, i.e., the meaning inferred from the same word may vary across resolutions or input transformations, which ultimately harming downstream STR.
To address this issue, we introduce the Multi-scale Linguistic Alignment (MLA) module, which enforces appearance-invariant and scale-stable semantics by directly aligning the encoder-level \texttt{[CLS]} representations from all training views.

\textbf{Semantic motivation.}
Text images differ from natural images in that: characters must preserve identity regardless of resizing or stylistic perturbations; local masking should not alter the underlying lexical meaning; global shape deformation (handled by NSP) must still map to consistent textual semantics.
Thus, aligning the \texttt{[CLS]} token across views forces the encoder to retain a linguistic anchor that is robust to multi-scale visual variations.

Concretely, we take the \texttt{[CLS]} tokens from three views: the small-scale augmentation $\mathcal{E}(s_k)^{cls}$, the large-scale augmentation $\mathcal{E}(s_{k+1})^{cls}$, and the masked zoom-in view $\mathcal{E}(s_{masked})^{cls}$.
These views differ in resolution, appearance, and local completeness, making them suitable anchors for enforcing semantic consistency.
If the masked view were aligned only to $s_k$, the encoder would be biased toward coarse, resolution-invariant semantics and might underuse high-frequency details.
Aligning only to $s_{k+1}$ would overemphasize fine-scale appearance and weaken robustness to strong downsampling.
Using both as anchors provides symmetric supervision across the scale range, encouraging a single, scale-stable semantic representation.

We adopt an MSE loss to enforce this appearance-invariant alignment:
\begin{align}
\label{eq:align}
\mathcal{L}_{\text{mla}}
&= \left\|\mathcal{E}(s_{masked})^{cls}-\mathcal{E}(s_k)^{cls}\right\|_2^2 \\
&\quad + \left\|\mathcal{E}(s_{masked})^{cls}-\mathcal{E}(s_{k+1})^{cls}\right\|_2^2.
\end{align}
Operating purely at the encoder level avoids shortcut leakage between branches, while jointly regularizing semantics across coarse, fine, and masked views.

\textbf{Overall objective.}
Our final training objective is the sum of all three components:
\begin{equation}
\mathcal{L}
=\mathcal{L}_{\text{nsp}}+\mathcal{L}_{\text{mim}}+\mathcal{L}_{\text{mla}}.
\end{equation}
NSP supervises cross-scale structural consistency, MIM enforces local completion under occlusion, and MLA stabilizes linguistic representation across views, yielding a scale-aware and semantically coherent encoder for STR.
\section{Experiments}
\label{sec:experiment}
We evaluate MNSP in terms of overall recognition performance, component-wise effectiveness, and learned representation properties. Specifically, we first compare with prior methods on Union14M and six standard STR benchmarks, then conduct ablations on the proposed designs, and finally analyze the scale robustness and prediction targets of MNSP.

\subsection{Datasets}
\textbf{Pre-training Data.}
To enable scalable self-supervised learning, we perform pre-training on the large-scale unlabeled text image datasets, Union14M-U. It is a 10-million-image subset of Union14M~\cite{union14m_sstr_iccv23}, containing diverse unlabeled real-world text instances, which is collected with an IoU Voting method from 3 large datasets, \textit{i.e.}, Book32, OpenImage and Conceptual Captions(CC).

\textbf{Fine-tuning Data.}
For text recognition, we fine-tune our model on the large-scale labeled dataset Union14M-L. which is the labeled subset of Union14M~\cite{union14m_sstr_iccv23} consisting of 3.2 million samples. 

\textbf{Benchmarks.} 
We evaluate on six widely used datasets: three for regular text (\textit{i.e.}, IIIT5K~\cite{iiit_5k_bmvc12}, IC13~\cite{ic13_icdar13}, SVT~\cite{svt_iccv11}) and three for irregular text (\textit{i.e.}, IC15~\cite{icdar15_icdar15}, SVTP~\cite{svtp_iccv13}, CUTE80~\cite{cute80_eswa14}). Recent works have re-verified annotations to address mislabeling issues~\cite{union14m_sstr_iccv23,xiaomeng_spl24}. And the Union14M benchmark includes 0.41 million samples spanning seven challenging scenarios. Abbreviations for the subsets are defined as follows: Cur (Curve), M-O (Multi-Oriented), Art (Artistic), Ctx (Contextless), Sal (Salient), M-W (Multi-Words), and Gen (General).

\setlength{\tabcolsep}{2pt}
\begin{table*}[h]
    \small 
  \centering
      \caption{Performance on the English Union14M benchmark~\cite{union14m_sstr_iccv23}. Unless otherwise specified, all text recognizers are trained using real data from Union14M-L. MNSP$_{20ep}$ refers to pretraining for 20 epoches. }
  \begin{tabular}{ll|ccccccc|c|c}
    \toprule
    Method &Publisher  &Cur &M-O &Art &Ctx &Sal &M-W &Gen &Avg &Params\\
    \midrule
    ABINet~\cite{abinet_cvpr21} &\textit{CVPR} 21   &75.0 &61.5 &65.3 &71.1 &72.9 &59.1 &79.4 &69.2  &37M \\
    VisionLAN~\cite{visionlan_iccv21} &\textit{ICCV} 21   &70.7 &57.2 &56.7 &63.8 &67.6 &47.3 &74.2 &62.5  &33M \\
    MATRN~\cite{na2022matrn} &\textit{ECCV} 22   &80.5 &64.7 &71.1 &74.8 &79.4 &67.6 &77.9 &74.6  &44M \\
    PARSeq~\cite{parseq_eccv22} &\textit{ECCV} 22   &79.8 &79.2 &67.4 &77.4 &77.0 &76.9 &80.6 &76.9  &24M \\
    MIM~\cite{mae_cvpr22} &\textit{CVPR} 22  &84.8 &79.3 &71.1 &81.5 &81.0 &82.5 &82.0 &80.3  &36M \\
    DiG-S~\cite{dig_sstr_mm22} &\textit{MM} 22  &85.9 &83.5 &77.4 &82.5 &84.3 &84.0 &83.8 &83.0  & 36M \\
    MAERec-S~\cite{union14m_sstr_iccv23} &\textit{ICCV} 23  &81.4 &71.4 &72.0 &82.0 &78.5 &82.4 &82.5 &78.6  &36M \\
    
    SSM-S~\cite{ssm_ijcai24} &\textit{IJCAI} 24   &87.5 &85.8 &78.4 &\textbf{84.8} &85.2 &85.0 &84.0 &84.3  &36M \\
    LMIM~\cite{zhang2025lmim} &\textit{CVPR} 25  &87.5 &84.5 &\textbf{79.8} &84.3 &86.6 &86.1 &84.4 &84.7  &36M  \\
    \midrule
    NSP (Ours) &- &88.6 &85.6 &75.6 &82.9 &85.4 &86.1 &83.8 &84.0  &36M  \\
    MNSP (Ours) &- &88.9 &88.1 &78.3 &83.4 &86.9 &\textbf{87.2} &84.7 &85.4 &36M \\
    MNSP$_{20ep}$ (Ours) &- &\textbf{90.8} &\textbf{90.4} &78.4 &84.5 &\textbf{87.3} &86.5 &\textbf{85.6} &\textbf{86.2}  &36M  \\
    
    \bottomrule
  \end{tabular}

  \label{tab:union14m}
\end{table*}

\subsection{Implementation Details}
\textbf{Pre-training Phase}
We adopt ViT-Small (ViT-S)~\cite{vit_iclr21} encoder with 12 transformer layers as the default backbone. Input images are resized to $32 \times 128$, and divided into 256 patches, with each patch size of 4 × 4, with 80\% of the patches randomly masked in the MIM stage. The NSP and MIM modules share the encoder, and use the two-layers SA-CA-FFN decoders with an output feature dimensionality of 384 separately. Pre-training is conducted for 10 epochs by default, including a one-epoch warm-up. We use the AdamW optimizer with a cosine learning rate schedule, starting from a base learning rate of 3e-4, and a batch size of 512.

\textbf{Fine-tuning Phase}
For downstream STR tasks, the pre-trained encoder is combined with a six-layer transformer decoder, following a standard encoder-decoder architecture with 512-dimensional hidden states~\cite{dig_sstr_mm22}. We adopt data augmentation strategies consistent with ABINet~\cite{abinet_cvpr21}. Fine-tuning is performed using the AdamW optimizer with a learning rate of 1e-4 and a batch size of 512. The learning rate scale across layers is set to 1 by default. We train for 10 epochs and the maximum sequence length is set to 25 for both Union14M benchmark and six common STR benchmarks.

\setlength{\tabcolsep}{2pt}
\begin{table}
  \small
  \centering
    \caption{Results on six common STR benchmarks~\cite{iiit_5k_bmvc12,ic13_icdar13,svt_iccv11,icdar15_icdar15,svtp_iccv13,cute80_eswa14}.  FT-data refers to the fine-tuning dataset. U14M-L is the abbreviation of Union14M-L. The symbol $^*$ denotes results re-calculated using the original results from those works. The symbol $^\dag$ denotes using corrected label. }
  \begin{tabular}{l|cccccc|c}
    \toprule
    Method & IIIT5K &IC13 & SVT &IC15 &SVTP &CUTE &Avg \\
    \midrule
    ABINet~\cite{abinet_cvpr21} &97.2 &97.2 &95.7 &87.6 &92.1 &94.4 &94.0  \\
    VisionLAN~\cite{visionlan_iccv21} &96.3 &95.1 &91.3 &83.6 &85.4 &92.4 &91.3 \\
    MATRN~\cite{na2022matrn} &98.2 &97.9 &96.9 &88.2 &94.1 &97.9 &95.5   \\
    PARSeq~\cite{parseq_eccv22} &98.0 &96.8 &95.2 &85.2 &90.5 &96.5 &93.8   \\
    MaskOCR-S~\cite{maskocr_tmlr24} &98.0 &97.8 &96.9 &90.2 &94.9 &96.2 &95.6  \\
    DiG-S~\cite{dig_sstr_mm22} &98.7 &97.8 &98.5 &88.9 &92.7 &96.5 &95.5  \\
    CCD~\cite{ccd_sstr_iccv23}  &98.0 &98.3 &96.4 &90.3 &92.7 &98.3 &95.6$^*$  \\
    MAERec-S~\cite{union14m_sstr_iccv23} &98.0 &97.6 &96.8 &87.1 &93.2 &97.9 &95.1  \\
    SSM-S~\cite{ssm_ijcai24} & 99.0 & 98.3 &\textbf{97.8} &89.5 &94.0 &98.3 &96.1  \\
    LMIM~\cite{zhang2025lmim}  &98.5 &98.0 &97.7 &88.7 &94.0 &\textbf{98.6} &95.9  \\
        \midrule
    NSP (Ours) & 98.2 & 98.5 & 96.3 & 89.3 & 92.7 & 97.2 & 95.4  \\
    NSP$^\dag$  (Ours) & 98.8 & \textbf{98.6} & 96.9 & 90.8 & 93.3 & 97.6 & 96.0 \\
    MNSP (Ours) & 98.7 & 98.4 & 96.6 & 90.5 & 94.3 & 97.9 & 96.1  \\
    MNSP$^\dag$ (Ours) & \textbf{99.3} & 98.5 & 97.2 & \textbf{91.4} & \textbf{95.2} & 98.3 & \textbf{96.7}  \\
    \bottomrule
  \end{tabular}

  \label{tab:sixbench}
  
\end{table}

\subsection{Performance Comparison}
We comprehensively evaluate our MNSP framework across a diverse set of benchmarks, including the large-scale Union14M benchmark and six widely adopted standard STR benchmarks. The results consistently demonstrate the superiority of MNSP over previous state-of-the-art methods in both accuracy and generalization capability.

\textbf{Union14M Benchmark.}
Table~\ref{tab:union14m} shows that \textbf{MNSP} achieves the highest Avg (\textbf{85.4\%}) on Union14M. Notably, we achieve a state-of-the-art average accuracy of \textbf{86.2\%} when pre-training for 20 epochs. The gains are most pronounced in scale-sensitive categories (e.g., Curve and Multi-Oriented, where text deformations are inherently more susceptible to scale variations) while remaining competitive elsewhere. This indicates that explicit next-scale pre-training delivers stronger recognition under severe scale and geometric variations with a comparable model size and training protocol.

\textbf{Six Common STR Benchmarks.}
We evaluate our method using corrected labels on the six common benchmarks. As summarized in Table~\ref{tab:sixbench}, \textbf{MNSP} achieves the best Avg of \textbf{96.7\%}. It remains competitive on the easier sets (IIIT5K / IC13 / SVT) and shows clear gains on the more challenging IC15 / SVTP / CUTE, indicating strong, scale-robust generalization without task-specific tuning.

\subsection{Ablation Study}
For fast turn-around, we pre-train on two Union14M-U subsets (Book32 and OpenImages) and fine-tune for 5 epochs; results are reported on the seven Union14M scenarios (Curve, M-O, Artistic, Contextless, Salient, M-W, General). 

\begin{figure}[h]
\centering
\includegraphics[width=\columnwidth]{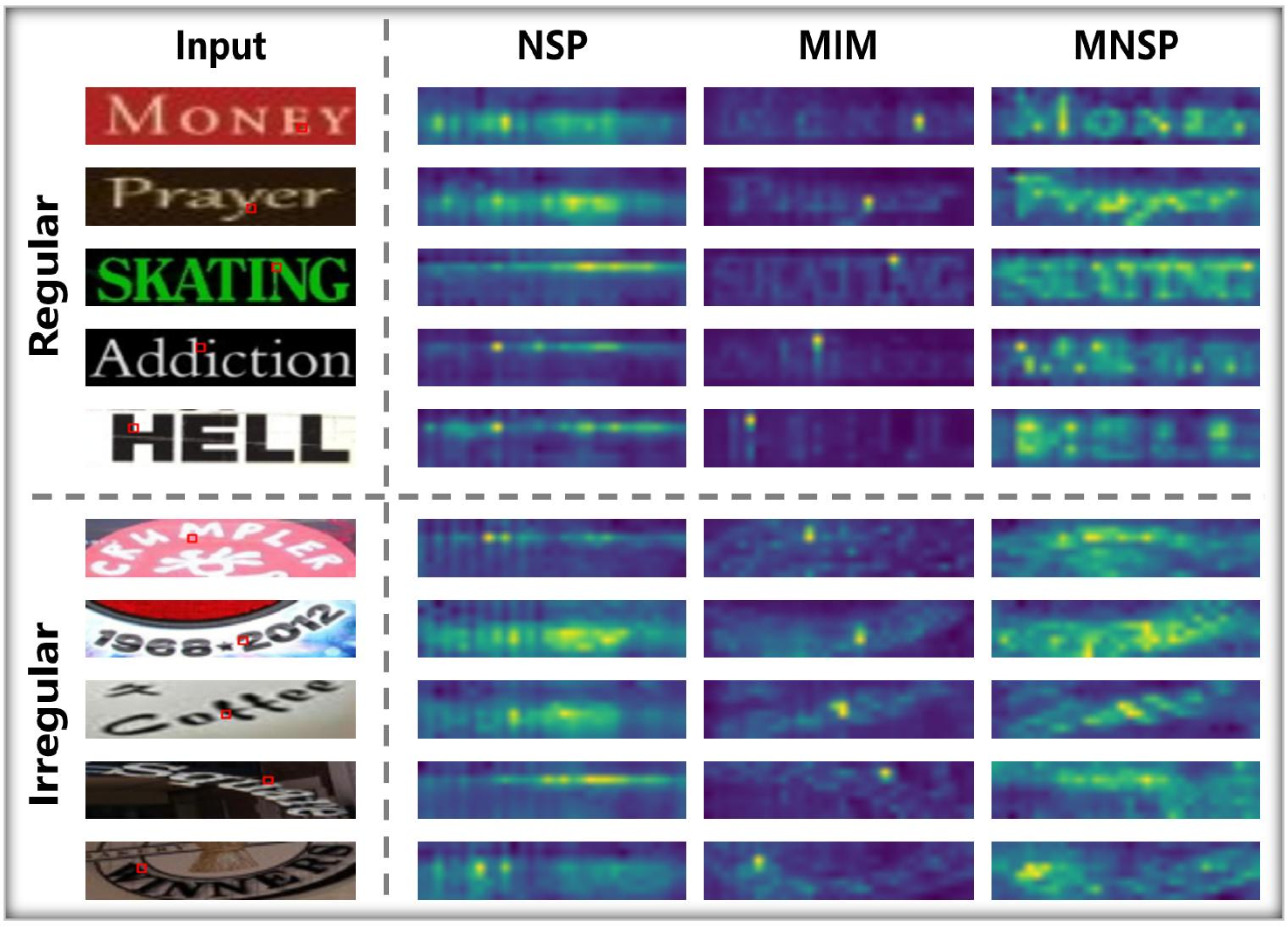}
\caption[]{Query-based attention visualization on the last encoder layer. Red boxes mark query locations. MNSP attends coherently over the whole text region; MIM is highly local; NSP is overly diffuse, and these patterns remain consistent on both regular and irregular text images.
}
\label{fig:complement}
\end{figure}

\textbf{Complementarity between NSP and MIM.}
Table~\ref{tab:complement} quantitatively validates the non-redundant contributions of NSP and MIM. To uncover the underlying mechanism, Figure~\ref{fig:complement} visualizes the token-level attention distributions. In the standalone \textbf{NSP} paradigm, all tokens remain globally visible to predict higher-resolution features. This lack of spatial bottleneck causes the encoder's attention to over-diffuse into irrelevant background regions. Conversely, \textbf{MIM} forces masked tokens to heavily depend on adjacent unmasked patches, resulting in strictly localized, short-ranged attention that completely loses the global layout. 

\begin{table}[h]
\small
  \centering
  \caption{Complementarity between NSP and MIM. NSP+MIM  is our MNSP method.}
  \begin{tabular}{l|ccccccc|c}
    \toprule
    Component   &Cur &M-O &Art &Ctx &Sal &M-W &Gen &Avg \\
    \midrule
    NSP & 83.8 & 78.5 & 72.1 & 80.6 & 81.1 & 81.9 & 81.3 & 79.9 \\
    MIM & 83.7 &78.2 &72.3 &80.3 &81.0 &82.5 &81.7 &80.0 \\
    NSP+MIM & \textbf{84.6} & \textbf{80.3} & \textbf{72.9} & \textbf{82.4} & \textbf{82.5} & \textbf{83.7} & \textbf{82.1} & \textbf{81.2} \\
        \bottomrule
  \end{tabular}
  \label{tab:complement}
\end{table}

Our \textbf{MNSP} naturally neutralizes these two extremes through mechanism duality. The strict local reconstruction constraint of MIM prevents NSP's global attention from diverging into the background, forcing text tokens to selectively attend to structurally similar tokens across the entire text region. This mutual regularization endows the encoder with both global layout awareness and local focus, which explains the consistent accuracy gains in Table~\ref{tab:complement}, especially on highly deformed irregular texts that require precise multi-granular perception.


\textbf{Multi-scale Linguistic Alignment and Guidance.}
Note that the ablations in Table~\ref{tab:linguistic} are performed by removing specific components from the full MNSP framework. The results show that enabling either \emph{MLA} or \emph{guidance} improves performance, and using both is best. Their effectiveness reflects complementary roles in tightening branch coupling: Guidance injects appearance-invariant cues from augmented views to stabilize semantics, while the MLA further regularizes cross-view/scale representations and reduces drift. Together they better couple the branches and decorrelate errors, yielding the largest improvement on \textbf{Avg}.

\textbf{Effectiveness of Data Preprocessing.}
Similarly, Table~\ref{tab:data} evaluates our preprocessing strategies on the full framework. Both \emph{zoom-in} (for MIM) and \emph{data augmentation} (for NSP) help, and the combination performs best. The gains are intuitive: zoom-in amplifies local detail and makes masked reconstruction harder, yielding stronger supervision for fine strokes; data augmentation prevents inter-branch information leakage (shortcut learning) and injects appearance-invariant linguistic cues from different renderings of the same text. Together they reduce both local ambiguity and semantic drift, producing the largest overall improvement.

\begin{table}[t]
  \centering
\small
  \caption{Effectiveness of multi-scale linguistic alignment and guidance.}
  \begin{tabular}{lc|ccccccc|c}
    \toprule
    MLA & Guide & Cur & M-O & Art & Ctx & Sal & M-W & 
    Gen & Avg \\
    \midrule
    - & - & 84.6 & 80.1 & 71.6 & 82.3 & 82.1 & 81.9 & 82.0 & 80.6 \\
    - & $\checkmark$ & 83.9 & 79.8 & 73.2 & 81.8 & \textbf{82.7} & 83.2 & 81.9 & 80.9 \\
    $\checkmark$ & - & \textbf{84.8} & 79.8 & 72.8 & 80.7 & 82.4 & \textbf{84.1} & 82.1 & 81.0 \\
    $\checkmark$ & $\checkmark$ & 84.6 & \textbf{80.3} & \textbf{72.9} & \textbf{82.4} & 82.5 & 83.7 & \textbf{82.1} & \textbf{81.2} \\
    \bottomrule
  \end{tabular}

  \label{tab:linguistic}
\end{table}


\begin{table}[t]
\small
  \centering
  \caption{Effectiveness of data preprocessing.}
  \begin{tabular}{lc|ccccccc|c}
    \toprule
    Zoom-in & Aug. & Cur & M-O & Art & Ctx & Sal & M-W & 
    Gen & Avg \\
    \midrule
    - & - & 84.1 & 79.1 & 71.8 & 80.1 & 82.0 & 83.6 & 81.7 & 80.3 \\
    - & $\checkmark$ & 84.0 &79.5 &74.3 &80.6 &80.7 &83.4 &81.8 &80.6 \\
    $\checkmark$ & - & \textbf{85.2} &\textbf{80.5} &72.1 &80.5 &82.1 &83.2 &81.9 &80.8 \\
    $\checkmark$ & $\checkmark$ & 84.6 & 80.3 & \textbf{72.9} & \textbf{82.4} & \textbf{82.5} & \textbf{83.7} & \textbf{82.1} & \textbf{81.2} \\
    \bottomrule
  \end{tabular}
  \label{tab:data}
\end{table}

\begin{figure*}[!t]
\centering
\includegraphics[width=0.99\linewidth]{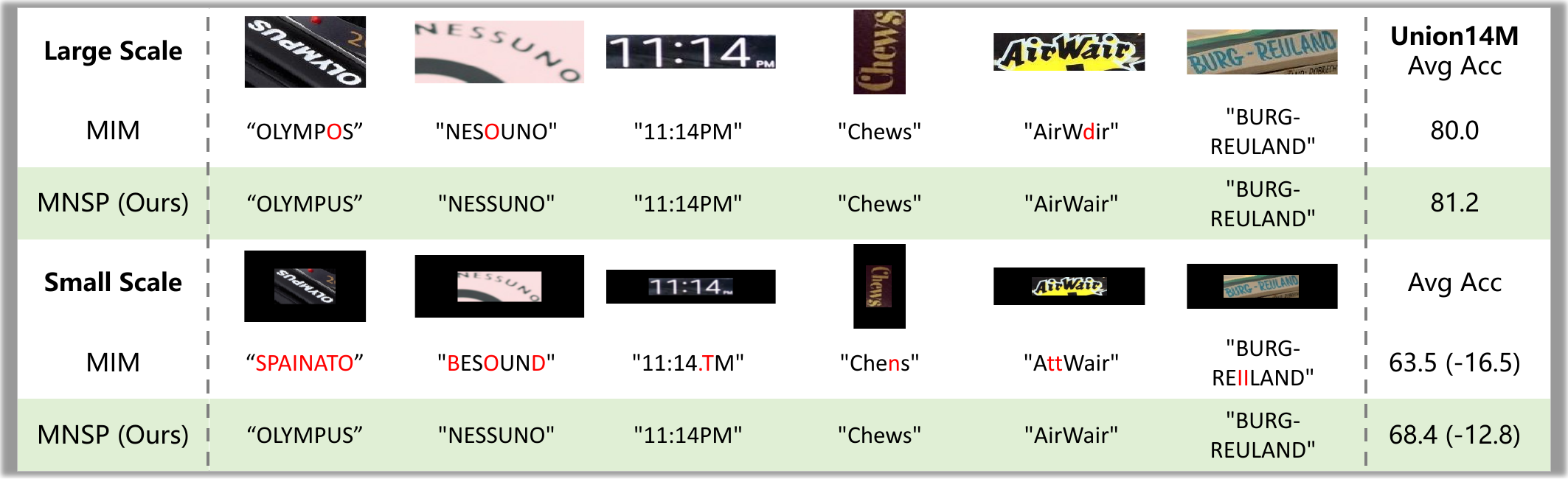}
\caption[]{Qualitative and quantitative comparison under controlled scale shift. Top: fine-tuned and evaluated on standard (large-scale) Union14M. Bottom: fine-tuned and evaluated on the small-scale corpus created by 0.5× shrink + black padding. MNSP remains correct in both regimes, while MIM often fails at small scale; the average accuracy drop is markedly smaller for MNSP (-12.8 vs. -16.5).}
\label{fig:scale}
\end{figure*}

\textbf{Scale Robustness of MNSP.}
Due to the fixed input size required by the model, all images are resized to a common resolution, which inevitably introduces scale variation. To systematically evaluate scale robustness, we design two complementary experiments: grouping by original scale and controlling relative text size.
Table~\ref{tab:scale} evaluates robustness under an original-size split: all images are sorted by original size and split into small and large halves, then resized to $32{\times}128$ at inference.
Larger originals experience stronger downsampling (stroke thinning/aliasing), causing accuracy drops for both methods. \textbf{MNSP} degrades notably less than \textbf{MIM} (a smaller Avg drop and a higher score on the large group), indicating that explicit next-scale pretraining builds a scale bridge that preserves global layout while retaining recoverable stroke cues after resizing.


Figure~\ref{fig:scale} confirms the same trend under a complementary, controlled protocol: we synthesize a small-scale corpus via $0.5{\times}$ shrink + black padding and compare against the standard (large-scale) setting. Qualitatively, \textbf{MNSP} stays correct across both regimes while \textbf{MIM} exhibits small-scale errors (letter substitutions/splits). Quantitatively, the average accuracy drop under this synthetic shift is markedly smaller for \textbf{MNSP} than for \textbf{MIM}. Together with Table~\ref{tab:scale}, these results show that multi-scale pretraining encourages scale-consistent features that maintain both word-line continuity and fine strokes across substantial changes in character size.

\textbf{Effect of Prediction Targets.}
Table~\ref{tab:target} demonstrates that predicting \emph{fixed features} from a frozen teacher significantly outperforms raw \emph{pixel} reconstruction. Pixel-level targets force the model to overfit low-level textures and are highly sensitive to cross-scale resampling artifacts. In contrast, fixed feature targets provide a stable, semantically rich supervision signal. This ensures the encoder learns robust, appearance-invariant structural representations rather than trivial image statistics.

\begin{table}[t]
\small
  \centering
  \caption{Robustness of MNSP in different scale groups.}
  \begin{tabular}{lc|ccccccc|c}
    \toprule
    Model & Scale & Cur & M-O & Art & Ctx & Sal & M-W & 
    Gen & Avg \\
    \midrule
    MIM & small & 86.4 &82.2 &71.6 &82.3 &84.2 &83.3 &83.4 &81.9 \\
    MNSP & small & 86.7 &83.2 &70.9 &83.0 &84.7 &82.6 &83.3 &82.1 \\
    \midrule
    MIM & large & 80.7 &73.9 &72.9 &77.9 &77.6 &81.7 &80.2 &77.8{\footnotesize (-4.1)}  \\
    MNSP & large & 82.6 &79.2 &72.4 &81.2 &81.4 &82.9 &81.6 &80.2{\footnotesize (-1.9)} \\
    \bottomrule
  \end{tabular}

  \label{tab:scale}
\end{table}

\begin{table}[t]
  \centering
    \caption{Effect of different prediction targets.}
  \begin{tabular}{l|ccccccc|c}
    \toprule
    Targets   &Cur. &M-O &Art. &Ctl. &Sal. &M-W &Gen. &Avg \\
    \midrule
    Pixel  & 83.8 & 78.4 & 71.0 & 80.4 & 80.0 & 83.0 & 81.5 & 79.7 \\
    Feature  & \textbf{84.6} & \textbf{80.3} & \textbf{72.9} & \textbf{82.4} & \textbf{82.5} & \textbf{83.7} & \textbf{82.1} & \textbf{81.2} \\
        \bottomrule
  \end{tabular}
  \label{tab:target}
\end{table}
\section{Conclusion}
\label{sec:conclusion}
In this paper, we proposed Masked Next-Scale Prediction (MNSP) to explicitly address the intrinsic multi-scale challenge in self-supervised Scene Text Recognition. While we introduced Next-Scale Prediction (NSP) to capture the hierarchical evolution from coarse layouts to fine strokes, we identified a critical flaw: its unconstrained nature leads to severe attention diffusion. The core innovation of MNSP lies in synergistically coupling NSP with Masked Image Modeling (MIM) to exploit their mechanism duality. MIM's strict local reconstruction constraints successfully neutralize NSP's global attention diffusion, while NSP provides explicit cross-scale structural guidance to prevent MIM from making blind local guesses. Complemented by a Multi-scale Linguistic Alignment (MLA) module, these two paradigms are deeply coupled to ensure both structural and semantic consistency. Extensive experiments validate this design, demonstrating state-of-the-art performance and exceptional robustness against severe scale and layout variations. Ultimately, MNSP establishes that integrating explicit multi-scale priors with local fine-grained perception is a superior self-supervised paradigm, paving a new path for robust visual text analysis and beyond.

\section*{Acknowledgments}
This work was supported by the National Natural Science Foundation of China under Grants U24B6012, 62406167, 62376266, and 62406318.

{
    \small
    \bibliographystyle{ieeenat_fullname}
    \bibliography{main}
}


\end{document}